\documentclass[letterpaper]{article} % DO NOT CHANGE THIS
\usepackage{aaai24}  % DO NOT CHANGE THIS
\usepackage{times}  % DO NOT CHANGE THIS
\usepackage{helvet}  % DO NOT CHANGE THIS
\usepackage{courier}  % DO NOT CHANGE THIS
\usepackage[hyphens]{url}  % DO NOT CHANGE THIS
\usepackage{graphicx} % DO NOT CHANGE THIS
\usepackage{natbib}  % DO NOT CHANGE THIS AND DO NOT ADD ANY OPTIONS TO IT
\usepackage{caption} % DO NOT CHANGE THIS AND DO NOT ADD ANY OPTIONS TO IT
\usepackage{algorithm}
\usepackage{algorithmic}

\usepackage{newfloat}
\usepackage{listings}

\usepackage[dvipsnames]{xcolor}

\usepackage{amssymb} % For additional symbols
\usepackage{cite} % For proper citation formatting
\usepackage{amsmath} 
\usepackage{makecell}
\usepackage{amssymb}
\usepackage{enumitem}
\usepackage{colortbl}
\definecolor{lightgray}{gray}{0.90} % 90% white
\usepackage{xcolor}
\usepackage{arydshln}
\usepackage{graphicx}
\usepackage{array}
\usepackage{xcolor}
\usepackage{booktabs}
\usepackage{tabularx}
\usepackage{makecell}
\usepackage{pifont}
\nocopyright
\newcommand{\cmark}{\textcolor{green}{\ding{51}}}%
\newcommand{\xmark}{\textcolor{red}{\ding{55}}}%
\DeclareUnicodeCharacter{FF0C}{,}

\usepackage{newfloat}
\usepackage{listings}
\DeclareCaptionStyle{ruled}{labelfont=normalfont,labelsep=colon,strut=off} % DO NOT CHANGE THIS
\floatstyle{ruled}
\newfloat{listing}{tb}{lst}{}
\floatname{listing}{Listing}
\title{CROPS: A Deployable Crop Management System Over All Possible State
Availabilities}
\author{
    Jing Wu\equalcontrib \thanks{jingwu6@illinois.edu}，
    Zhixin Lai \equalcontrib,
    Shengjie Liu \equalcontrib,
    Suiyao Chen \equalcontrib,\\
    Ran Tao,
    Pan Zhao,
    Chuyuan Tao,
    Yikun Cheng,
    Naira Hovakimyan
}
\usepackage{bibentry}
\begin{document}

\maketitle

\begin{abstract}
Exploring the optimal management strategy for nitrogen and irrigation has a significant impact on crop yield, economic profit, and the environment. To tackle this optimization challenge, this paper introduces a deployable \textbf{CR}op Management system \textbf{O}ver all \textbf{P}ossible \textbf{S}tate availabilities (CROPS). CROPS employs a language model (LM) as a reinforcement learning (RL) agent to explore optimal management strategies within the Decision Support System for Agrotechnology Transfer (DSSAT) crop simulations. A distinguishing feature of this system is that the states used for decision-making are partially observed through random masking. Consequently, the RL agent is tasked with two primary objectives: optimizing management policies and inferring masked states. This approach significantly enhances the RL agent's robustness and adaptability across various real-world agricultural scenarios. Extensive experiments on maize crops in Florida, USA, and Zaragoza, Spain, validate the effectiveness of CROPS. Not only did CROPS achieve State-of-the-Art (SOTA) results across various evaluation metrics such as production, profit, and sustainability, but the trained management policies are also immediately deployable in over of ten millions of real-world contexts. Furthermore, the pre-trained policies possess a noise resilience property, which enables them to minimize potential sensor biases, ensuring robustness and generalizability. Finally, unlike previous methods, the strength of CROPS lies in its unified and elegant structure, which eliminates the need for pre-defined states or multi-stage training. These advancements highlight the potential of CROPS in revolutionizing agricultural practices.

\end{abstract}

\section{Introduction}
Food security is a crucial goal in contemporary agriculture, highlighting the significance of key management practices such as nitrogen fertilization and water irrigation. These techniques are essential not only for increasing crop yields and ensuring a stable food supply but also play a vital role in sustaining environmental health.  Traditional best practices in these domains, informed by empirical experience and scholarly research, are now being tested against the backdrop of changing climatic and market conditions. This raises concerns about their continued effectiveness, underscoring the need for more innovative, efficient, and adaptable management systems. Such systems are essential for developing strategies that are responsive to varying conditions and aimed at specific goals, including economic profitability. This study seeks to contribute to this need by applying advanced AI methods to enhance agricultural practices, addressing these significant challenges in the pursuit of more sustainable and productive farming methodologies.

\begin{table}[t]
\centering
\small
\setlength{\tabcolsep}{2pt} % Adjust column separation
\resizebox{0.35\textwidth}{!}{ % Adjust the width to 80% of the text width
    \begin{tabular}{lcccc}
    \toprule
    \textbf{Criterias} & \makecell{\textbf{MLP-based} \\ \textbf{Agent}}  & \makecell{\textbf{IM-based} \\ \textbf{Agent}} & \makecell{\textbf{LM-based} \\ \textbf{Agent}} & \makecell{\textbf{CROPS} \\ \textbf{Agent (Ours)}} \\
    \midrule
    \makecell[l]{\textbf{Unified} \\ \textbf{Framework}} & \cmark & \xmark & \cmark & \cmark \\ \hdashline
    \makecell[l]{\textbf{Readily} \\ \textbf{Deployable}} & \xmark & \cmark & \xmark & \cmark \\ \hdashline
    \makecell[l]{\textbf{Noise} \\ \textbf{Resilience}} & \xmark & \xmark & \xmark & \cmark \\
    \hdashline
    \makecell[l]{\textbf{Deployable} \\ \textbf{Scenarios}} & \textcolor{red}{0} & \textcolor{red}{1} & \textcolor{red}{0} & \textcolor{green}{$10^8$} \\
    \bottomrule
    \end{tabular}
}
\caption{Overview of the critical properties of the proposed method compared with previous SoTA methods.}
\label{tab:compares}
\vspace{-4mm}
\end{table}

\begin{figure*}[t]
  \centering
  \includegraphics[width=0.75\linewidth]{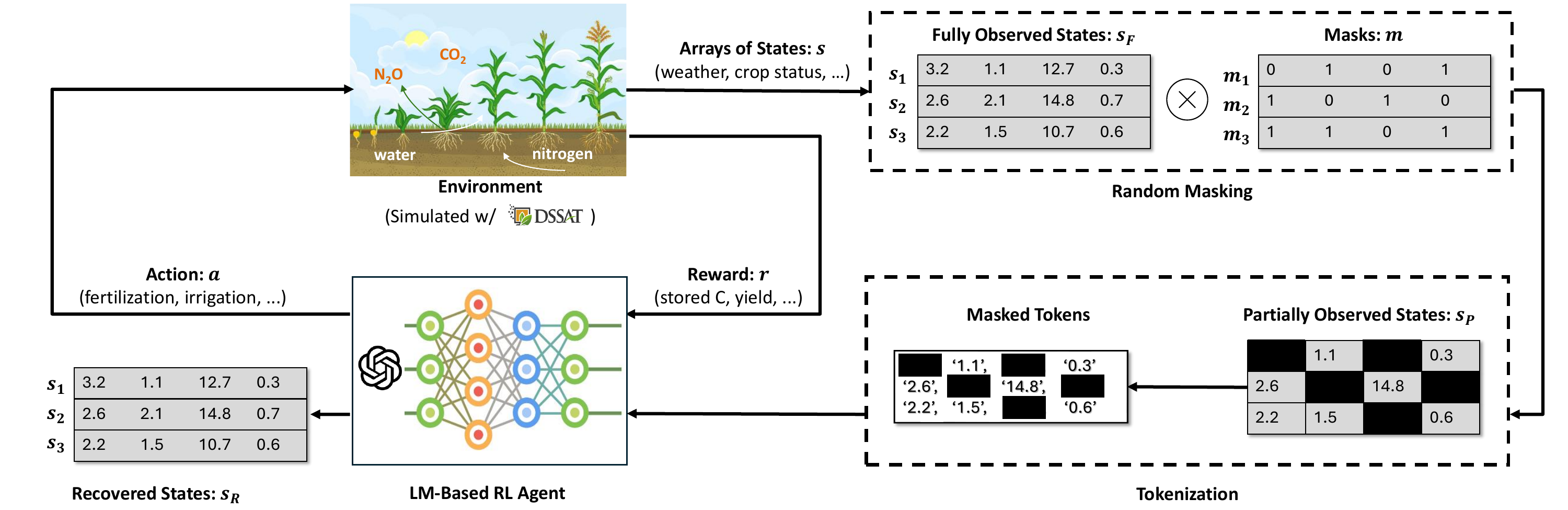}
   \caption{Framework and pipeline of the intelligent crop management system using LM-based RL}
   \label{fig:Framework}
   \vspace{-5mm}
\end{figure*}

Recent advancements in agricultural technology have introduced multi-layer perception (MLP)-based reinforcement learning (RL) agents (MLP-based Agents) and language-model-based reinforcement learning agents (LM-based Agents) for training nitrogen (N) and irrigation management policies using the Gym-DSSAT simulator \cite{romain2022gym, wu2022optimizing, wu2024new, wu2024exploratory}. Their research demonstrated the ability of these policies to surpass a baseline by producing higher yields or achieving similar yields with less N input under full observation conditions. However, the practical implementation of these policies in real-world scenarios is hindered by their reliance on comprehensive observational data, such as nitrate leaching and plant N uptake, which are typically not readily available to farmers. Addressing this gap, a recent study presents an intelligent crop management framework that adeptly combines reinforcement learning (RL), imitation learning (IL), and crop simulations using DSSAT and Gym-DSSAT \cite{tao2022optimizing}. In this paper, we refer to their trained agent as the imitation learning-based agent (IM-based Agent). This approach enhances the adaptability and applicability of the management policies to real-world agricultural settings by effectively addressing the challenge of partial observation.

While IL has proven effective in refining existing agricultural strategies by better aligning them with the practical realities of farming, it's crucial to recognize the variability in the availability of states in real-world scenarios. This variation is often case-specific, dictated by factors such as the deployment of sensors and the unique characteristics of different environments. Consequently, one state that is observable and accessible in one location may not be available in another, posing a significant challenge. This inconsistency in state availability can severely limit the applicability of a pre-trained RL agent from imitation learning.

Additionally, the two-stage training approach used in RL and IL represents a significant limitation in the context of agricultural management optimization. Unlike an integrated, end-to-end framework, this method typically involves using an expert policy, pretrained in a fully observed setting, to guide the RL agent in scenarios with only partial observations. Such a bifurcated training process can potentially lead to suboptimal optimization of crucial resources like nitrogen and water. This is because the prior knowledge in expert policy, developed under the assumption of complete information, may not be transferred effectively to settings where only limited data is available. Consequently, this approach might result in management strategies that are both less efficient and less effective.

Addressing the aforementioned challenges, our research pivots to developing a more robust and universally applicable RL agent trained within a unified framework. Prior studies have suggested the utilization of language models (LMs) as enhanced RL agents, demonstrating state-of-the-art (SoTA) performance across diverse scenarios. Although potent in their capabilities, these models are primarily configured for scenarios with full access to all states in simulations, limiting their direct deployability in real-world settings. Building on this limitation, we introduce the masking technique as a crucial auxiliary component in our optimization task.

To be more specific, we propose an intelligent crop management framework that incorporates a powerful LM-based RL agent, state masking strategy, and crop simulations via Gym-DSSAT. We illustrate the overall framework in Figure ~\ref{fig:Framework}. We transitioned from a traditional MLP-based RL agent to a more powerful LM-based agent, exhibiting an improved ability to enhance crop yields and promote sustainability amidst the complexities of optimization tasks.  More importantly, we have also implemented a state masking strategy that replicates the inherent uncertainties of real-world agricultural scenarios. Consequently, the LM-based RL agent is charged with a twofold task: executing management decisions and reconstructing obscured states. This development not only enables the RL agent to make smarter decisions when information is incomplete but also strengthens its ability to make reliable and noise-agnostic decisions in the face of the unpredictability that farmers often face. In summary, the primary contributions of this work are as follows:

\begin{itemize}
\item 
The study delves into a pivotal but underexplored question: can LMs, utilizing random state masking and reconstruction, function as superior bi-task RL agents for crop management and missing state recovery?
\item 
We propose CROPS, the first, elegant and unified framework that is readily deployable, noise-resilient, and applicable to ten millions of real-world contexts as shown in Table~\ref{tab:compares}.

\item 
We empirically demonstrate that CROPS outperforms existing state-of-the-art approaches in extensive experiments, assessing metrics such as crop yield, resource utilization, environmental impact, and, most critically, robustness in both fully observed and partially observed settings.
\end{itemize}

\section{Related Work}

\subsection{Crop Management with RL and Crop Models}
RL is increasingly being applied in the domain of crop management, aiming to enhance decision-making through simulation-based strategies. Early experiments, such as wheat management in France \cite{garcia1999rl-wheat} and irrigation optimization for maize in Texas \cite{sun2017rl-irrigation}, encountered limitations due to restrictive state and action spaces. To overcome these limitations, subsequent studies have broadened the scope of RL applications, incorporating more complex scenarios and a variety of crop models \cite{wu2022optimizing, kallenberg2023nitrogen, madondo2023swat}. The integration of RL with sophisticated crop simulation models like APSIM and DSSAT has been facilitated by innovative platforms such as CropGym and Gym-DSSAT. These platforms enable dynamic, real-time decision-making within simulations \cite{brockman2016openai, overweg2021cropgym, romain2022gym}, offering detailed daily interactions between the RL agent and the simulated environment, which is a crucial advancement for optimizing key management practices such as nitrogen and irrigation management \cite{wu2022optimizing}. Despite these advancements, challenges remain, particularly in deploying trained RL agents in real-world settings. Previous efforts have utilized imitation learning to adapt strategies developed under fully observed conditions to partially observed settings, although the ability to generalize these strategies to different real-world scenarios remains limited \cite{tao2022optimizing}.

\subsection{Language Models as RL Agent}

Recently, foundation models have been applied across various domains, such as language \cite{liu2024llmeasyquant, lai2024adaptive, liu2024towards}, speech \cite{chen2024neural, sun2024contextual,yu2024stochastic}, computer vision \cite{chen2024bridging, li2024ecnet, lai2024residual, wu2023genco, wu2023extended,wang2024balanced}, medical \cite{liu2024timemattersexaminetemporal, chen2024deep, chen2020optimal, chen2018data, chen2019claims}, geography \cite{yu2024harnessing}, robotics \cite{gao2023autonomous, gao2024decentralized, gao2024adaptive, zhang2020manipulator}and health \cite{xu2024machine}. In particular, foundation models have increasingly been applied to RL tasks that require language processing, such as natural language instructions \cite{garg2022lisa}. In a complementary approach, RL trajectories are encoded into token sequences, which foundation models then process to serve as inputs for deep RL architectures \cite{li2022pre}. Building on these developments, some recent methodologies conceptualize RL as a sequence modeling problem, employing foundation models directly to predict future states or actions \cite{janner2021offline,lee2022multi,huang2022language,wu2024new}. These innovative strategies have demonstrated remarkable success, significantly enhancing control schemes across a variety of robots and agents \cite{huang2022language,huang2022inner,raman2022planning,mees2023grounding,liu2024towards,chen2023open,ahn2022can,liang2023code}. Beyond these stages, previous studies have also tested the viability of language models as autonomous RL agents in agricultural management \cite{wu2024new}. 

Although these trials were successful, they stopped short of actual field deployment. Addressing this gap, this research pioneers the deployment of LMs for formulating optimal management strategies in real-world agricultural settings, where access to complete state information is often challenging.

\subsection{Masking Strategy for Improved Robustness}
Masked language modeling and its autoregressive counterparts, such as BERT \cite{devlin2018bert} and GPT \cite{radford2018improving, radford2019language, brown2020language}, have proven to be highly effective methods for pre-training in natural language processing (NLP) \cite{zhang2024reconsidering}. Building on this success, the concept of a masking strategy has been extended to the domain of computer vision and cross-modal, yielding significant benefits \cite{gong2022reverse, chen2023point, Sun_2024_CVPR, fu2024detecting, he2022masked, li2023scaling}. Similarly, this approach has been applied to graph neural networks \cite{liu2024graphsnapshot, li-etal-2023-joyful,schlichtkrull2020interpreting}. Beyond the modalities of vision and language, the masking technique has also been successfully implemented in the representation learning of tabular datasets and other domains\cite{xin2024let,yoon2020vime,wu2024switchtab,chen2023recontab,wu2023hallucination, shi2017combining}. To the best of our knowledge, this study represents the first effort to leverage the masking strategy to enhance the robustness of RL agents and enable the deployment of adaptable policies under varying state availability.

\section{Method}
In this section, we formally introduce CROPS. The first sub-section outlines how the crop management process can be formulated as a Markov Decision Process (MDP). The second sub-section details the masking strategy employed within the crop management setting during both training and inference phases. Finally, the last sub-section presents the unified framework that integrates a masking strategy with language models.

% This innovative approach not only enhances the robustness of the pre-trained policy but also increases its adaptability, making it suitable for deployment across various real-world scenarios where the state is partially observed.

\subsection{Problem Formulation}
\label{sec: formulation}
Following established paradigms \cite{tao2022optimizing,wu2022optimizing,wu2024new}, we formulate nitrogen fertilization and irrigation management as a finite MDP. Specifically, we use $t$ to denote a day. For each day, $s_t$ represents the state on that day. The state $s_t$ includes key data pertaining to weather, plant growth, and soil conditions, including root depth and cumulative nitrate levels, as observed in the simulation for that day. Given the environmental state $s_t$, RL agents are trained to select an action $a_t$ from the action space $\mathcal{A}$. This selection is guided by a policy $\pi(s_t,\theta_t)$, where $\theta_t$ represents the policy parameters on that particular day. Notably, a pre-trained language model is employed to represent the policy. For the action $a_t$, it comprises two key decisions: the quantity of nitrogen fertilizer, denoted as $N_t$, and the amount of irrigation water, $W_t$, to be applied. The effectiveness of these decisions is quantified by the reward $r_t(s_t,a_t)$, calculated based on the outcomes of $s_t$ and $a_t$. The reward function is defined as follows:
\begin{align}
\label{eq:reward}
    &\text{if harvest occurs at } t: \nonumber \\
    % &\quad r_t(s_t, a_t) = w_1 Y - w_2 N_t - w_3 W_t - w_4 N_{l,t}  \nonumber\\
        &\quad r_t(s_t, a_t) = w_1 Y - w_2 N_t - w_3 W_t - w_4 N_{l,t} , \nonumber\\
    &\text{otherwise:}  \\
    &\quad r_t(s_t, a_t) = - w_2 N_t - w_3 W_t - w_4 N_{l,t}  ,\nonumber
\end{align}
where $w_1, w_2, w_3, w_4$ represent four custom weight factors, $Y$ denotes the yield at harvest, and $N_{l,t}$ indicates the amount of nitrate leaching on a given day, respectively.

Both $Y$ and $N_{l,t}$ are derived from the state variable $s_t$. The reward function design, characterized by the weights $w_1, w_2, w_3, w_4$, plays a crucial role in guiding the agent's strategy. The agent's objective is to determine the optimal policy $\pi(s_t,\theta_t)$, which selects action $a_t$ to maximize the total future discounted return. This return, defined as $R_t = \sum_{\tau=t}^T \gamma^{\tau-t} r_\tau$, represents the accumulated rewards from the current action $a_t$ to future rewards, each discounted by the factor $\gamma$.

\subsection{Mimicking Real-World Observations through State Masking}
\label{sec: masking}
We innovatively utilize a masking strategy to mimic the states that can be accessed in reality, preparing the trained RL agent for deployment and ensuring stable performance. For the training stage, we depict the masking process in Figure~\ref{fig:Framework}. For a batch of states, we define their original and fully observed condition as $s_{F}$. For each state, we sample a subset of its features and mask out the selected ones using masks denoted by $m$. More specifically, $m$ consists of a series of zeros and ones, where the ones correspond to the state features we retain, and the zeros correspond to the features selected for masking following a uniform distribution. The masked ratio $\alpha$ is defined as the ratio of the number of masked states to the total number of states. Then, the masked and partially observed state is defined as $s_{P} = s_{F} \odot m $, where $\odot$ denotes the element-wise operation between the fully observed state $s_{F}$ and the mask $m$. The operation strategy is straightforward: When the element in the mask $m$ is 0, we replace the corresponding element in  $s_{P}$ with ``\#".

% we randomly sample features, replacing the selected elements in $m$ with the string $\text{\#}$, following a uniform distribution. We refer to this as ``random sampling". 

In the inference stage of real-world applications, fully observed states $s_{F}$ are no longer available. Instead, all state features are partially observed due to real-world constraints such as the availability of sensors, weather conditions, or financial limitations. Consequently, during deployment, all states will naturally be partially observed. Therefore, the RL agent will directly utilize these partially observed states $s_{P}$ directly for decision-making.

In the fields of computer vision and NLP, masking strategies typically require high masking ratios due to the redundancy and structured nature of images and texts. In contrast, our approach adopts a lower masking ratio, reflecting the less redundant and less structured nature of the states we analyze. This allows a modest masking ratio, such as 30\%, to effectively create a challenging pre-task, prompting the RL agent to infer missing features and learn latent dependencies. One the other hand, using a higher masking ratio in this context could disrupt training stability. To mitigate potential center bias and enhance adaptability, we sample $\alpha$ uniformly within a specified range. This approach allows the trained RL) agent to perform effectively across diverse real-world applications. By reducing its reliance on specific state features, the agent can make informed decisions under varying conditions of state availability, thereby fostering more robust and noise-resistant decision-making.

\begin{table}[t]
\centering
\resizebox{0.35\textwidth}{!}{ % Adjust the width to 80% of text width
    \begin{tabular}{l r r r r l }
    \toprule
    & \makecell{$w_1$\\ ($Y$)}  & \makecell{$w_2$\\ ($N_t$)} & \makecell{$w_3$ \\($W_t$)} & \makecell{$w_4$ \\($N_{l,t}$)} & Reward's Focus \\ \midrule
    RF 1 & 0.158 & 0.79    & 1.1     & 0 & Economic profit \\ 
    RF 2 & 0.158 & 0.79    & 0       & 0 & Free water \\ 
    RF 3 & 0.158 & 0       & 1.1     & 0 & Free N fertilizer \\ 
    RF 4 & 0.158 & 1.58    & 1.1     & 0 & Doubled N price \\ 
    \bottomrule
    \end{tabular}
}
\caption{Weights used in each reward function (RF) defined by equation \eqref{eq:reward}.}
\label{table:weight}
\vspace{-5mm}
\end{table}

\subsection{Policy Training with RL and Masking Strategy}
\label{sec: framework}
In this study, we employ the Deep Q-Network (DQN) framework from \cite{mnih2015human-deepRL} to train our agent. The objective is to learn an optimal policy that maximizes the future discounted return, denoted by $R_t$. Within this framework, we utilize a LM to predict the action-value function, i.e., the Q-function. More specifically, we define this Q-function as $Q^\pi(s,a) = \mathbb{E}[R_t|s_t=s,a_t=a,\pi]$. We use this Q-function to estimate the expected future return from the current state $s$ and action $a$.

The fundamental difference between the proposed framework and the current framework is that the LM serves as a bi-task RL agent. Specifically, LM not only estimates the Q-values but is also designed to recover the masked or missing states. On one hand, the optimization goal is to refine the parameters of the Q-network to explore the optimal Q function, $Q^\star(s,a)$, which represents the highest possible return given the current state $s$ and action $a$. For decision-making, we employ a greedy policy defined as $a_t^\star = \max_{a\in \mathcal{A}}Q^\star(s_t,a)$. On the other hand, the input state for the Q function is partially observed due to our designed masking strategy. Therefore, the Q function accepts the input state $s_{P}$, i.e., $Q^\pi(s_{P},a) = \mathbb{E}[R_t|s_t=s_{},a_t=a,\pi]$. The language model also plays the role of a transition function, $T(s_{p}, a) = \hat{s}_{F}$, to recover the partially observed states to the approximated fully observed ones. In summary, we define the overall framework, which effectively explores the optimal policy and recovers masked states using the following loss function \( L_i(\theta_i) \) as follows:
\begin{equation}
L_i(\theta_i) \triangleq L_{i,1}(\theta_i) + \lambda L_{i,2}(s_{F}, \hat{s}_{F}),
\end{equation}
where
\begin{equation}
L_{i,1}(\theta_i) \! \triangleq
  \!\! \mathop{\!\mathbb{E}}_{(s_{F},a,r,s')}\!\left[r\!+\!\gamma \max_{a'\in \mathcal{A}}Q(s'\!,a';\theta_i^{-}) \!-\! Q(s_{F},a;\theta_i)\!\right]\!
\end{equation}
and
\begin{equation}
L_{i,2}(s_{F}, \hat{s}_{F}) = \mathrm{MSE}(s_{F}, \hat{s}_{F}).
\end{equation}
Here, $s_{P}$, $s_{F}$, $\hat{s}_{F}$, $a$, $r$, and $s'$ denote the partially observed state, fully observed state, recovered fully observed state, action, reward, and next state, respectively. $\lambda$ is designed to balance the the two optimization objectives. Additionally, $\gamma$ represents the discount factor, $\theta_i^{-}$ denotes the parameters of a previously defined target network, and $\mathrm{MSE}$ stands for mean squared error. The tuples $(s_{F},a,r,s')$ for the loss function are randomly sampled from the replay buffer, a collection of prior state-action-reward-next state tuples accumulated during training.

\begin{table*}[ht]
\renewcommand{\arraystretch}{0.9}\addtolength{\tabcolsep}{0pt}
\small
\centering
\resizebox{0.6\textwidth}{!}{ % Adjust the width to 80% of text width
    \begin{tabular}{l rrrrrrr}
    \toprule
    Florida Case & \makecell{N Input \\ (kg/ha) $\downarrow$} & \makecell{Irrigation\\ (L/m$^2$) $\downarrow$} & \makecell{Yield\\ (kg/ha) $\uparrow$} & RF1 $\uparrow$ & RF2 $\uparrow$ & RF3 $\uparrow$ & RF4 $\uparrow$  \\ 
    \midrule
    Empirical Baseline & 360 & 394 & 10772 & 984 & 1417 & 1269 & 700 \\  
    \midrule
    Policy1: Traditional Agent & {200} & \textbf{120} & 10852 & {1425} & 1557 & 1538 & 1267 \\ 
    Policy1: LM-based Agent & \textbf{122} & {192} & {11402} & {1464} & {1675} & {1590} & {1337} \\  
    \rowcolor{lightgray} Policy1: CROPS Agent (Ours) & {200} & {186} & \textbf{12281} & \textbf{1578} & \textbf{1782} & \textbf{1736} & \textbf{1420} \\  
    \midrule
    Policy2: Traditional Agent & 200 & 732 & 11244 & 813 & {1619} & 971 & 655 \\ 
    Policy2: LM-based Agent & \textbf{160} & \textbf{510} & {11474} & \textbf{1126} & {1687} & {1252} & {999} \\  
    \rowcolor{lightgray} Policy2: CROPS Agent (Ours) & \textbf{160} & {762} & \textbf{13417} & {1155} & \textbf{1993} & \textbf{1281} & \textbf{1028} \\  
    \midrule
    Policy3: Traditional Agent & 19920 & \textbf{108} & 10865 & -1.4e4 & -1.4e4 & {1598} & {-3.0e4} \\  
    Policy3: LM-based Agent & \textbf{10000} & {264} & {13152} & {-6.1e3} & \textbf{-5.8e3} & {1788} & {-1.4e4} \\  
    \rowcolor{lightgray} Policy3: CROPS Agent (Ours) & {10120} & {144} & \textbf{13789} & \textbf{-5.9e3} & \textbf{-5.8e3} & \textbf{2020} & \textbf{-1.3e3} \\  
    \midrule
    Policy4: Traditional Agent & 160 & 102 & \textbf{10358} & 1398 & \textbf{1510} & 1524 & {1272} \\ 
    Policy4: LM-based Agent & {160} & {36} & {10192} & {1428} & {1468} & {1555} & {1302} \\  
    \rowcolor{lightgray} Policy4: CROPS Agent (Ours) & \textbf{160} & \textbf{36} & {10161} & \textbf{1439} & {1479} & \textbf{1565} & \textbf{1313} \\  
    \bottomrule
    \end{tabular}
}
\caption{Comparison of different policies in the Florida case study. The best value is highlighted in \textbf{bold}. Details of each reward function can be found in Table \ref{table:weight}.}
\label{table:florida}
\vspace{-4mm}
\end{table*}

% \caption{The evaluation results of our trained policies, comparing them with previous SoTA methods and baseline policies. `Policy x' refers to the policy optimized using the reward function `RF x'. The `RF x' column details the cumulative rewards for each policy, calculated in accordance with `RF x'. Details of each reward function can be found in Table \ref{table:weight}. The best value is highlighted in \textbf{bold}.}
% \label{table:TP_results}
%    \vspace{-3mm}
% \end{table*}

\section{Experiments and Results}
In this section, we introduce the initial experimental setup for the subsequent experiments, including the datasets and settings. Following this, we provide the details of the training and evaluation processes. We then present the evaluation results, where the performance of the proposed method is compared against SoTA approaches in both fully observed and partially observed settings. Additionally, we include critical ablation studies to further analyze our method's effectiveness.

\subsection{Experimental Setup}

The studies examining training policies for nitrogen and irrigation management in maize crops encompassed two separate case studies, both employing real-world data. The initial case study took place in a simulated setting modeled after Florida, USA, in 1982, whereas the second was based on simulations of Zaragoza, Spain, in 1995. 

For a more comprehensive evaluation of the proposed framework, DQN was utilized to train the RL agent using a masking strategy for both partially and fully observed states. The performance of all developed policies was benchmarked against existing SoTA methods. Specifically, the baseline for the Florida study was drawn from a maize production guide for Florida farmers \cite{wright2022field}, while the baseline for the Zaragoza study was based on survey data regarding maize farming practices in Zaragoza \cite{malik2019dssat,skhiri2012impact}.

\begin{table*}[ht]
\small
\centering
\resizebox{0.6\textwidth}{!}{ % Adjust the width to 80% of text width
    \begin{tabular}{lrrrrrrr}
    \toprule
    Zaragoza Case & \makecell{N Input \\ (kg/ha) $\downarrow$} & \makecell{Irrigation\\ (L/m$^2$) $\downarrow$} & \makecell{Yield\\ (kg/ha) $\uparrow$} & RF1 $\uparrow$ & RF2 $\uparrow$ & RF3 $\uparrow$ & RF4 $\uparrow$ \\ 
    \midrule
    Empirical Baseline & 250 & 752 & 10990 & 712 & 1539 & 909 & 514 \\  
    \midrule
    Policy1: Traditional Agent & {240} & {330} & 10477 & {1103} & 1466 & 1292 & 913 \\ 
    Policy1: LM-based Agent & \textbf{160} & {354} & {10806} & {1192} & \textbf{1581} & {1318} & {1065} \\  
    \rowcolor{lightgray} Policy1: CROPS Agent (Ours) & {200} & \textbf{210} & \textbf{10989} & \textbf{1347} & {1578} & \textbf{1505} & \textbf{1189} \\  
    \midrule
    Policy2: Traditional Agent & 200 & 1068 & 10923 & 393 & {1568} & 551 & 235 \\ 
    Policy2: LM-based Agent & \textbf{160} & {1032} & 10856 & \textbf{453} & \textbf{1588} & {580} & \textbf{327} \\  
    \rowcolor{lightgray} Policy2: CROPS Agent (Ours) & {240} & \textbf{1021} & \textbf{10989} & {423} & {1546} & \textbf{613} & 234 \\  
    \midrule
    Policy3: Traditional Agent & 10640 & \textbf{324} & 10626 & -7083 & -6727 & {1323} & {-1.5e4} \\  
    Policy3: LM-based Agent & \textbf{10000} & {342} & {10903} & {-6553} & {-6177} & {1347} & \textbf{-1.4e4} \\  
    \rowcolor{lightgray} Policy3: CROPS Agent (Ours) & \textbf{10000} & {338} & \textbf{10989} & \textbf{-6535} & \textbf{-6163} & \textbf{1364} & \textbf{-1.4e4} \\  
    \midrule
    Policy4: Traditional Agent & 120 & \textbf{336} & {9601} & 1053 & 1422 & 1147 & {958} \\ 
    Policy4: LM-based Agent & \textbf{160} & 348 & {10250} & \textbf{1110} & {1493} & {1268} & {984} \\  
    \rowcolor{lightgray} Policy4: CROPS Agent (Ours) & \textbf{160} & 396 & \textbf{10742} & {1135} & \textbf{1570} & \textbf{1261} & \textbf{1008} \\  
    \bottomrule
    \end{tabular}
}
\caption{Comparison of different policies in the Zaragoza case study. The best value is highlighted in \textbf{bold}. Details of each reward function can be found in Table \ref{table:weight}.}
\label{table:TP_results2}
\vspace{-5mm}
\end{table*}

\subsection{Evaluation Metrics and Implementation Details}

\subsubsection{Evaluation Metrics.} The framework was implemented to train the RL agent under conditions of both partial and full observation. In these settings, the method involved testing with four different reward functions, each designed to showcase the adaptability of the framework to various agricultural trade-offs. These include balancing crop yield, N fertilizer usage, irrigation water consumption, and environmental impacts. This diversity in reward functions enables the framework to be evaluated across a spectrum of scenarios and objectives, demonstrating its versatility in addressing different agricultural management challenges.

In each case study, we applied reward functions consistent with those described in prior research \cite{tao2022optimizing, wu2024new}. Specifically, four unique reward functions for $r_t$, derived from Equation \eqref{eq:reward}, were employed to train the RL agent. A single trained policy was selected for evaluation for each reward function, with detailed parameters for each listed in Table \ref{table:weight}.

RF1 measures the economic profit (\$/ha) accrued by farmers, calculated based on the prevailing market prices of maize and the costs associated with N fertilizer and irrigation water, as referenced from \cite{mandrini2022exploring} and \cite{wright2022field}. RF2-RF4 explore variations of economic profit under different hypothetical scenarios: RF2 assumes irrigation water is free; RF3 assumes N fertilizer is free; and RF4 models a scenario where the price of N fertilizer is doubled.

% Unlike RF1-RF4, which focus primarily on economic profit, RF5 incorporates an environmental dimension, specifically targeting nitrate leaching. Nitrate leaching is an environmental concern due to its contribution to issues like eutrophication of water bodies and soil degradation \cite{di2002nitrate}. RF5 is designed to balance yield, N fertilizer, and irrigation usage, while placing a significant emphasis on reducing nitrate leaching, thus aiming to minimize environmental impact while still ensuring favorable economic returns.

\subsubsection{Implementation Details.} 
The RL agent in the study employs a combination of DistilBERT and a three-layer fully connected neural network for feature adaptation. The process begins with DistilBERT encoding the state inputs into 768-dimensional embeddings. Notably, the parameters of DistilBERT are trained end-to-end in this model. After this initial encoding, the embeddings are passed through fully connected layers, one with 512 units and the other with 256 units. The final layer in this sequence is responsible for mapping these processed embeddings to the action space, completing the flow from the input state to the actionable output in the RL framework. The discrete action space is defined as follows:

\begin{align}
\label{eq:action space}
    \mathcal A =\{40k  \frac{\textrm{kg}}{\textrm{ha}} \text{ N fertilizer }\; \&\; 6k \frac{\textrm{L}}{m^2} \textrm{ Irrigation water} \},
\end{align}
where $k=0,1,2,3,4$, resulting in a total of 25 possible actions. This action space design incorporates standard quantities of N fertilizer and irrigation water that are typically applied by farmers in a single day. It also allows for a wide range of options, aiding the discovery of effective policies. The discount factor is meticulously set at 0.99. To facilitate the neural network's updates, Pytorch is employed alongside the Adam optimizer \cite{kingma2014adam}, characterized by an initial learning rate of 1e-5 and a batch size of 512. This setup is strategically chosen to optimize the learning process while ensuring efficient computation.

Applying DistilBERT's tokenizer to numerical values causes significant training instability due to multiple token splits, resulting in large variances for small numerical differences. For instance, 360 tokenizes into [9475], while 361 splits into [4029, 2487], leading to disproportionate representations and instability. Tokenizing decimals worsens this issue, as 0.1 translates into [1014, 1012, 1015], causing unnecessary token proliferation and inefficiency. To address this, we developed a preprocessing technique that normalizes numerical values to the range [0, 300] and uses only the integer part for tokenization. This ensures each number corresponds to a single token, simplifying and stabilizing the process. By focusing on integers, we reduce the token set to 27 distinct tokens, including 25 feature-specific tokens and two special tokens ([CLS] and [SEP]). This approach improves training stability and computational efficiency, essential for optimizing crop management using RL and language models.

We evaluate the masking ratio varies from 0\% to 100\%. Generally, larger masking ratios require longer training but demonstrate better generalization during deployment. We set $\lambda$ to 0.02 to balance these two optimization goals.

% \textcolor{red}{we use attention\_mask as LM input instead of replacing the original string with 0. attention\_mask consists of 0 and 1}

% \subsection{Results of Experiments}

\subsection{Policy Training with Full Observation and Random Masking.} 

DQN was implemented for training with all states available. However, we intentionally masked some of the states to enable the RL agent to better mimic real-world observations. We tested four different reward functions to demonstrate the adaptability of our framework to various trade-offs among crop yield, nitrogen fertilizer use, irrigation water use, and environmental impact.

The evaluation results of the trained policies are presented in Table~\ref{table:florida} and Table~\ref{table:TP_results2}. While the LM-based agent with random masking is not primarily designed to pursue SoTA results but rather to explore a more robust and deployable RL agent, it still outperforms previous SoTA methods and empirical baselines across most evaluation metrics (i.e., different RFs) and various  geographic locations, as a by-product. These consistent improvements across various reward functions that prioritize different optimization objectives underscore the agent’s adaptability in optimizing for diverse agricultural goals.

Notably, unlike previous work \cite{wu2024new}, which transforms states into descriptive language to enrich their semantic meaning, we found that direct tokenization of state variables can achieve similar results when using language models as agents. This indicates that language models can understand the underlying relationships between tokens and rewards without requiring redundant descriptions. Consequently, this approach is not only more straightforward to implement but also simplifies the preprocessing of states.

\begin{figure}[t]
  \centering
  \includegraphics[width=0.9\linewidth]{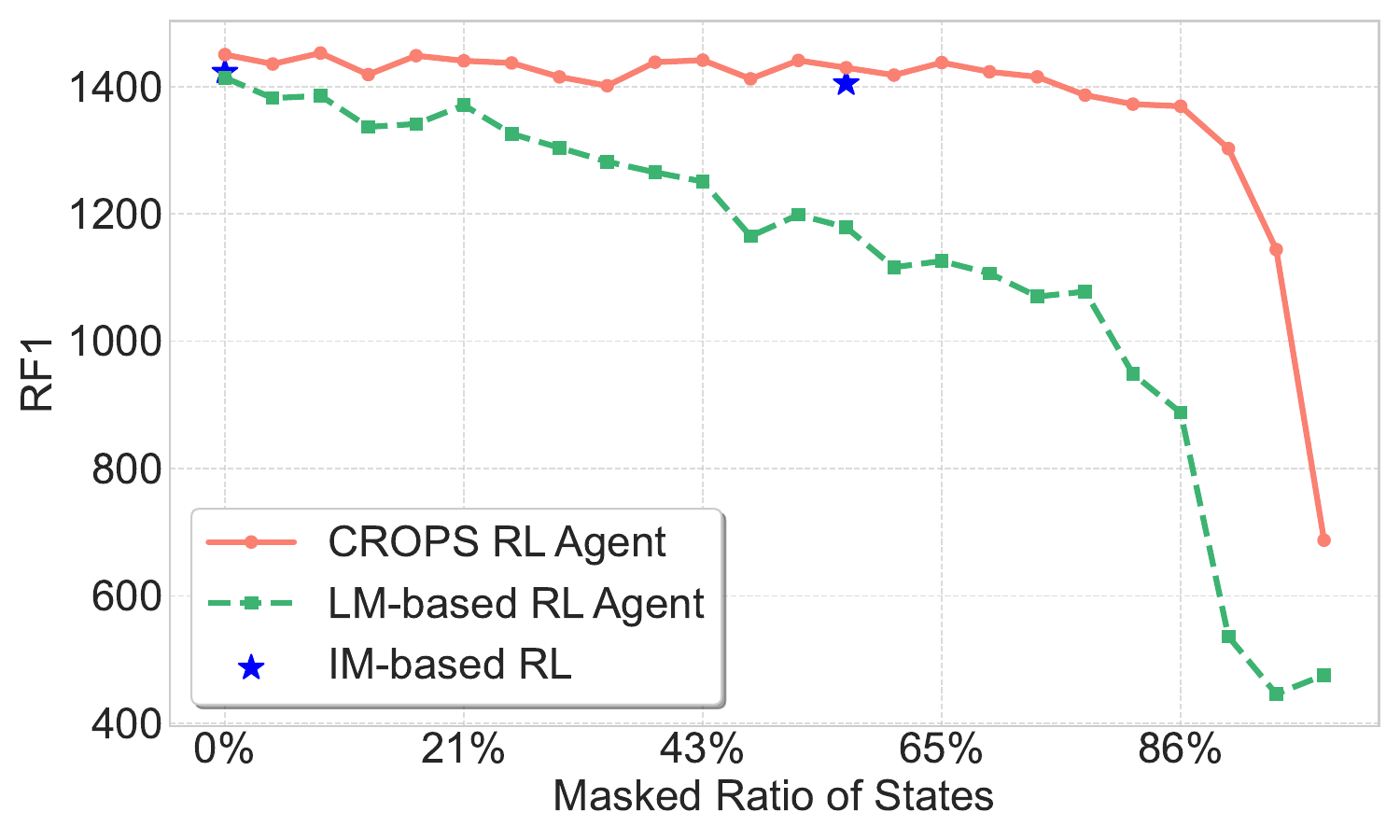}
   \caption{Performance degradation with the decrease in the availability of states.}
   \label{fig:partial}
      \vspace{-3mm}
\end{figure}

\subsection{Policy Evaluation under Partial Observation}
In the previous section, we included masked training with all states available from DSSAT. However, many of these states are not readily measurable or accessible to farmers due to limitations in available instruments in reality. Although there have been attempts to leverage imitation learning to guide partially observed agents in accomplishing crop management tasks, these approaches rely on predefined partially observed states \cite{tao2022optimizing}.

To address this issue, we pre-trained an LM-based RL agent with random masking. After the training, we evaluated the trained RL agent under partial observation settings. In this stage, we randomly masked a specific percentage of the states, defined as $\alpha$. For each $\alpha$, we kept its value unchanged but randomly selected masked states. We averaged the results of such experiments over 100 trials and reported the results in Figure~\ref{fig:partial}. Notably, $\alpha$ varies from 0\% to 100\% during inference and evaluation. As the available states gradually decrease, we observed a corresponding decrease in RF1. However, the decreasing curve is significantly more moderate than the one without masking, i.e., LM-based RL Agent, as shown in Figure~\ref{fig:partial}.

More importantly, we compared the performance of CROPS RL agent with IM-based RL agent \cite{tao2022optimizing} shown in blue stars in Figure~\ref{fig:partial}. PROPS not only surpassed the performance of the IM-based agent but also demonstrated significant advantages in real-world applicability. The Mask-based RL agent's adaptability to various state availabilities makes it highly deployable across diverse scenarios. These observations strongly support the state-agnostic nature of our method and its ease of deployment, highlighting its potential for broad and effective application.

\subsection{Ablation Studies}

\subsubsection{The Hyperparameter for Weighted Loss}
In this section, we conduct an important ablation study on the critical hyperparameter $\lambda$, which is designed to balance the optimization of state recovery and crop management tasks. The results are shown in Table~\ref{table:lambda}. While the optimal $\lambda$ is 0.02, this value may vary slightly based on different locations. However, the optimal range should remain on the scale of $10^{-2}$.

\begin{table}[t]
\centering
\resizebox{0.3\textwidth}{!}{ % Adjust the width to 80% of text width
    \begin{tabular}{l cccc }
    \toprule
    {$\lambda$} & 0 & 0.01 & \textbf{0.02}& 0.05\\ 
    \midrule
    Yield (kg/ha) $\uparrow$ & 11402 & 12138 & \textbf{12281} & 8961\\ 
    RF1 $\uparrow$ & 1464 & 1502 & \textbf{1578} & 1281\\ 
    \toprule
    $\lambda$ & 0.1 & 0.2 & 0.5 & 1 \\ 
    \midrule
    Yield $\uparrow$ & 11530 & 11760 & 10440 & 9949\\ 
    RF1 (kg/ha) $\uparrow$ & 1411 & 1475 & 862 & 817\\ 
    \bottomrule
    \end{tabular}
}
\vspace{-2mm}
\caption{Performance comparison with different values of $\lambda$. The best values are shown in \textbf{bold}.}
\label{table:lambda}

\end{table}

\begin{table}[t]
\centering
\resizebox{0.35\textwidth}{!}{ % Adjust the width to 80% of the text width
    \begin{tabular}{l cccc }
    \toprule
    % {Masked States} & 0-10 & \textbf{0-12} & 0-18 & 0-23 & 5-10 & 0-15\\ 
    {Masked States} & 0-10 & \textbf{0-12} & 0-18 & 0-23 \\ 
    \midrule
    Yield (kg/ha) $\uparrow$ & 12280 & \textbf{12281} & 11491 & 10923  \\ 
    RF1 $\uparrow$ & 1577 & \textbf{1578} & 1365 & 1276 \\ 
    \bottomrule
    \end{tabular}
}
\caption{Performance comparison with different values of masking range. The best values are shown in \textbf{bold}.}
\label{table:range}
\vspace{-4mm}
\end{table}

\subsubsection{Mask Ranges}
While masking states enhances the generalization capacity and robustness of the RL agent, excessive masking can result in information loss and training challenges. To determine the optimal masking range, we conducted experiments whose results are presented in Table~\ref{table:range}. Our findings indicate that the optimal masking range is between 0 and 12 states. Consequently, the optimal $\alpha$ for each sampling falls within the range of 0 to 0.48. When $\alpha=0$, all states are fully available. When $\alpha=0.48$, 12 out of 25 states are masked out.

% \subsubsection{Fixed Mask} \textcolor{red}{do we need?}.

% \subsection{Visualization}
\section{Path to Deploy}
The management policies trained in the DSSAT-simulated environment may not perform optimally in real-world conditions due to uncertainties in weather and discrepancies between the simulated crop models and actual cropping systems. This well-known issue, referred to as the sim-to-real gap \cite{zhao2020sim}, underscores the difficulties in transferring RL policies from simulation to real-world scenarios. We address this critical challenge in the following methods.

\subsection{Closing the Sim-To-Real Gap}
To enhance the robustness of our trained management policies against the challenges posed by the sim-to-real gap, previous methods incorporate domain and dynamics randomization techniques \cite{peng2018sim2real, zhao2020sim}. This approach involves introducing variations in critical model parameters and randomizing conditions during policy training to mimic the potential variances and noises encountered in real-world scenarios. These perturbations encourage the policies to become resilient to noises during deployment.

While the primary focus of our current work is to establish the Mask-based RL framework for crop management, we recognize the significance of ensuring the robustness of these policies in real-world scenarios. Consequently, we plan to explore this aspect in a future study.

% However, our findings indicate that such randomization is unnecessary. Due to the masking strategy, the proposed masking pre-training method inherently provides noise resilience during deployment. This characteristic significantly mitigates the sim-to-real gap. We further demonstrate the robustness of the trained agent in the following subsection.

\subsection{Policy Evaluation with Measurement Noises}
When deploying pre-trained policies in practice, farmers depend on observable states derived from weather forecasts and soil moisture measurements. However, these data sources are often prone to inaccuracies due to forecast errors and sensor limitations. To simulate this real-world scenario, we conducted experiments by retrieving the true state of the environment from the simulator and introducing random measurement noise to one or more key observable state variables.

The values for measurement noise were determined based on real-world accuracy data from weather forecasts and commonly used soil moisture meters available on the market. For each level of measurement noise introduced, we evaluated the policy 400 times and reported the decrease rate of RF1 in scenarios where no noise was applied. As demonstrated in the Table~\ref{table:noise}, CROPS exhibits a smaller decrease in performance and delivers more satisfactory and robust results compared to previous methods. These findings demonstrate that the masking pre-training method inherently provides noise resilience during deployment, benefiting from its strategic masking approach.

\begin{table}[t]

\centering
\small
\resizebox{1 \columnwidth}{!}{
\begin{tabular}{l r r r r}\toprule
\makecell{Variables} & Noises & IM-based Agent &LM-based Agent &\makecell{CROPS Agent}   \\ \midrule
% Empirical Baseline     & N/A           & N/A    & N/A    &N/A    \\ 
% No Noise               &N/A            &N/A   &N/A     &N/A \\
Soil water content     &-+0.02         & \textbf{0.0}   &\textbf{0.0}     &\textbf{0.0}   \\ 
Soil water content     &-+0.05         &0.19   &0.14     &\textbf{0.11} \\
Temperature            &-+1            &1.1   &1.3    &\textbf{0.8}   \\
Temperature            &-+2            &13.9   &11.9   &\textbf{9.7} \\
Solar Radiation        &-+2\%          &\textbf{0.0}   &\textbf{0.0}     &\textbf{0.0} \\
Solar Radiation        &-+10\%         &\textbf{0.0}   &\textbf{0.0}     &\textbf{0.0} \\
Rain Fall              &90 \% Acc. &3.5   &3.2   &\textbf{2.7}\\
Leaf Area Index        &-+10\%         &0.30   &0.40     &\textbf{0.22} \\
Leaf Area Index        &-+20\%         &0.50   &0.80     &\textbf{0.31} \\ 
\hdashline
Soil water content         &-+0.02           &    &      &  \\ 
+Temperature               &-+2              &    &      &  \\ 
+Solar Radiation           &-+2\%            &15.8    &15.2      &  \textbf{12.1}\\ 
+ Rain Fall                &90 \% Acc.           &    &      &\\ 
+ Leaf Area Index          &-+20\%           &    &      &\\ 
\bottomrule
\end{tabular} 
}
\caption{Comparison of decrease rate (\%) of different methods with Policy1 under noises. The decrease rate is calculated with respect to RF1, where no noise was applied.}
\label{table:noise}

\vspace{-5mm}
\end{table}

% \subsection{Deployment on a Real Farm}
% The management policies we have developed can be directly implemented on real farms with varying levels of sensor availability. To configure the crop model within DSSAT, it is necessary to collect soil and climate data specific to the farm for field tests, which will also be used for policy training. Once trained, the policies can make daily management decisions based on current soil and weather conditions, such as determining the optimal amounts of nitrogen fertilizers and water to apply. Farmers or decision-makers can then follow these recommendations to implement management practices. After the harvest, the performance of these management policies can be assessed by analyzing the amount of water and fertilizer used and the resulting crop yield. We are collaborating with academic agent XXX and industry partners XXX to conduct future field experiments to validate the efficacy of the proposed framework. (Note: Exact information is masked with XXX due to the double-blind policy.)

\section{Conclusion}
 In this paper, we propose CROPS, a unified, widely deployable, and noise-resilient crop management system. CROPS combines reinforcement learning, language models, and crop simulations to explore optimal management policies. In our experiments, we demonstrate the superior performance of CROPS using the maize crop in Florida, USA, and Zaragoza, Spain. More importantly, we show that the proposed method is readily deployable across ten millions of possible state availabilities. Compared to previous approaches with single and pre-defined state settings, CROPS has the capacity to generalize to countless scenarios worldwide. Moreover, its noise-resilient properties enable robust deployment in the real world, minimizing the influence of sensor bias. We believe that such advancements could significantly contribute to the evolution of agricultural technology, leading to more general, robust, and sustainable farming practices worldwide.

\bibliography{aaai24}

\begin{thebibliography}{73}
\providecommand{\natexlab}[1]{#1}

\bibitem[{Ahn et~al.(2022)Ahn, Brohan, Brown, Chebotar, Cortes, David, Finn, Fu, Gopalakrishnan, Hausman et~al.}]{ahn2022can}
Ahn, M.; Brohan, A.; Brown, N.; Chebotar, Y.; Cortes, O.; David, B.; Finn, C.; Fu, C.; Gopalakrishnan, K.; Hausman, K.; et~al. 2022.
\newblock Do as i can, not as i say: Grounding language in robotic affordances.
\newblock \emph{arXiv preprint arXiv:2204.01691}.

\bibitem[{Brockman et~al.(2016)Brockman, Cheung, Pettersson, Schneider, Schulman, Tang, and Zaremba}]{brockman2016openai}
Brockman, G.; Cheung, V.; Pettersson, L.; Schneider, J.; Schulman, J.; Tang, J.; and Zaremba, W. 2016.
\newblock Openai gym.
\newblock \emph{arXiv preprint arXiv:1606.01540}.

\bibitem[{Brown et~al.(2020)Brown, Mann, Ryder, Subbiah, Kaplan, Dhariwal, Neelakantan, Shyam, Sastry, Askell et~al.}]{brown2020language}
Brown, T.; Mann, B.; Ryder, N.; Subbiah, M.; Kaplan, J.~D.; Dhariwal, P.; Neelakantan, A.; Shyam, P.; Sastry, G.; Askell, A.; et~al. 2020.
\newblock Language models are few-shot learners.
\newblock \emph{Advances in neural information processing systems}, 33: 1877--1901.

\bibitem[{Chen et~al.(2023{\natexlab{a}})Chen, Xia, Ichter, Rao, Gopalakrishnan, Ryoo, Stone, and Kappler}]{chen2023open}
Chen, B.; Xia, F.; Ichter, B.; Rao, K.; Gopalakrishnan, K.; Ryoo, M.~S.; Stone, A.; and Kappler, D. 2023{\natexlab{a}}.
\newblock Open-vocabulary queryable scene representations for real world planning.
\newblock In \emph{2023 IEEE International Conference on Robotics and Automation (ICRA)}, 11509--11522. IEEE.

\bibitem[{Chen et~al.(2019)Chen, Kong, Sun, Meng, and Li}]{chen2019claims}
Chen, S.; Kong, N.; Sun, X.; Meng, H.; and Li, M. 2019.
\newblock Claims data-driven modeling of hospital time-to-readmission risk with latent heterogeneity.
\newblock \emph{Health care management science}, 22: 156--179.

\bibitem[{Chen et~al.(2024{\natexlab{a}})Chen, Liu, Li, Wu, and Yao}]{chen2024deep}
Chen, S.; Liu, X.; Li, Y.; Wu, J.; and Yao, H. 2024{\natexlab{a}}.
\newblock Deep Representation Learning for Multi-functional Degradation Modeling of Community-dwelling Aging Population.
\newblock \emph{arXiv preprint arXiv:2404.05613}.

\bibitem[{Chen et~al.(2018)Chen, Lu, Xiang, Lu, and Li}]{chen2018data}
Chen, S.; Lu, L.; Xiang, Y.; Lu, Q.; and Li, M. 2018.
\newblock A data heterogeneity modeling and quantification approach for field pre-assessment of chloride-induced corrosion in aging infrastructures.
\newblock \emph{Reliability Engineering \& System Safety}, 171: 123--135.

\bibitem[{Chen et~al.(2020)Chen, Lu, Zhang, and Li}]{chen2020optimal}
Chen, S.; Lu, L.; Zhang, Q.; and Li, M. 2020.
\newblock Optimal binomial reliability demonstration tests design under acceptance decision uncertainty.
\newblock \emph{Quality Engineering}, 32(3): 492--508.

\bibitem[{Chen et~al.(2023{\natexlab{b}})Chen, Wu, Hovakimyan, and Yao}]{chen2023recontab}
Chen, S.; Wu, J.; Hovakimyan, N.; and Yao, H. 2023{\natexlab{b}}.
\newblock Recontab: Regularized contrastive representation learning for tabular data.
\newblock \emph{arXiv preprint arXiv:2310.18541}.

\bibitem[{Chen et~al.(2024{\natexlab{b}})Chen, Wang, Khalilian-Gourtani, Yu, Dugan, Friedman, Doyle, Devinsky, Wang, and Flinker}]{chen2024neural}
Chen, X.; Wang, R.; Khalilian-Gourtani, A.; Yu, L.; Dugan, P.; Friedman, D.; Doyle, W.; Devinsky, O.; Wang, Y.; and Flinker, A. 2024{\natexlab{b}}.
\newblock A neural speech decoding framework leveraging deep learning and speech synthesis.
\newblock \emph{Nature Machine Intelligence}, 1--14.

\bibitem[{Chen et~al.(2024{\natexlab{c}})Chen, Jing, Li, and Li}]{chen2024bridging}
Chen, Z.; Jing, L.; Li, Y.; and Li, B. 2024{\natexlab{c}}.
\newblock Bridging the domain gap: Self-supervised 3d scene understanding with foundation models.
\newblock \emph{Advances in Neural Information Processing Systems}, 36.

\bibitem[{Chen et~al.(2023{\natexlab{c}})Chen, Li, Jing, Yang, and Li}]{chen2023point}
Chen, Z.; Li, Y.; Jing, L.; Yang, L.; and Li, B. 2023{\natexlab{c}}.
\newblock Point cloud self-supervised learning via 3d to multi-view masked autoencoder.
\newblock \emph{arXiv preprint arXiv:2311.10887}.

\bibitem[{Devlin et~al.(2018)Devlin, Chang, Lee, and Toutanova}]{devlin2018bert}
Devlin, J.; Chang, M.-W.; Lee, K.; and Toutanova, K. 2018.
\newblock Bert: Pre-training of deep bidirectional transformers for language understanding.
\newblock \emph{arXiv preprint arXiv:1810.04805}.

\bibitem[{Fu et~al.(2024)Fu, Wang, Xin, Zhou, Chen, Ge, Janies, and Zhang}]{fu2024detecting}
Fu, Z.; Wang, K.; Xin, W.; Zhou, L.; Chen, S.; Ge, Y.; Janies, D.; and Zhang, D. 2024.
\newblock Detecting Misinformation in Multimedia Content through Cross-Modal Entity Consistency: A Dual Learning Approach.

\bibitem[{Gao et~al.(2024)Gao, Aubert, Saldana, Danielson, and Fierro}]{gao2024decentralized}
Gao, L.; Aubert, K.; Saldana, D.; Danielson, C.; and Fierro, R. 2024.
\newblock Decentralized Adaptive Aerospace Transportation of Unknown Loads Using A Team of Robots.
\newblock \emph{arXiv preprint arXiv:2407.08084}.

\bibitem[{Gao et~al.(2023)Gao, Cordova, Danielson, and Fierro}]{gao2023autonomous}
Gao, L.; Cordova, G.; Danielson, C.; and Fierro, R. 2023.
\newblock Autonomous multi-robot servicing for spacecraft operation extension.
\newblock In \emph{2023 IEEE/RSJ International Conference on Intelligent Robots and Systems (IROS)}, 10729--10735. IEEE.

\bibitem[{Gao, Danielson, and Fierro(2024)}]{gao2024adaptive}
Gao, L.; Danielson, C.; and Fierro, R. 2024.
\newblock Adaptive Robot Detumbling of a Non-Rigid Satellite.
\newblock \emph{arXiv preprint arXiv:2407.17617}.

\bibitem[{Garcia(1999)}]{garcia1999rl-wheat}
Garcia, F. 1999.
\newblock Use of reinforcement learning and simulation to optimize wheat crop technical management.
\newblock In \emph{Proceedings of the International Congress on Modelling and Simulation}, 801--806.

\bibitem[{Garg et~al.(2022)Garg, Vaidyanath, Kim, Song, and Ermon}]{garg2022lisa}
Garg, D.; Vaidyanath, S.; Kim, K.; Song, J.; and Ermon, S. 2022.
\newblock LISA: Learning interpretable skill abstractions from language.
\newblock \emph{Advances in Neural Information Processing Systems}, 35: 21711--21724.

\bibitem[{Gong et~al.(2022)Gong, Yao, Li, Zhang, Liu, Lin, and Liu}]{gong2022reverse}
Gong, Y.; Yao, Y.; Li, Y.; Zhang, Y.; Liu, X.; Lin, X.; and Liu, S. 2022.
\newblock Reverse Engineering of Imperceptible Adversarial Image Perturbations.
\newblock In \emph{International Conference on Learning Representations}.

\bibitem[{He et~al.(2022)He, Chen, Xie, Li, Doll{\'a}r, and Girshick}]{he2022masked}
He, K.; Chen, X.; Xie, S.; Li, Y.; Doll{\'a}r, P.; and Girshick, R. 2022.
\newblock Masked autoencoders are scalable vision learners.
\newblock In \emph{Proceedings of the IEEE/CVF conference on computer vision and pattern recognition}, 16000--16009.

\bibitem[{Huang et~al.(2022{\natexlab{a}})Huang, Abbeel, Pathak, and Mordatch}]{huang2022language}
Huang, W.; Abbeel, P.; Pathak, D.; and Mordatch, I. 2022{\natexlab{a}}.
\newblock Language models as zero-shot planners: Extracting actionable knowledge for embodied agents.
\newblock In \emph{International Conference on Machine Learning}, 9118--9147. PMLR.

\bibitem[{Huang et~al.(2022{\natexlab{b}})Huang, Xia, Xiao, Chan, Liang, Florence, Zeng, Tompson, Mordatch, Chebotar et~al.}]{huang2022inner}
Huang, W.; Xia, F.; Xiao, T.; Chan, H.; Liang, J.; Florence, P.; Zeng, A.; Tompson, J.; Mordatch, I.; Chebotar, Y.; et~al. 2022{\natexlab{b}}.
\newblock Inner monologue: Embodied reasoning through planning with language models.
\newblock \emph{arXiv preprint arXiv:2207.05608}.

\bibitem[{Janner, Li, and Levine(2021)}]{janner2021offline}
Janner, M.; Li, Q.; and Levine, S. 2021.
\newblock Offline reinforcement learning as one big sequence modeling problem.
\newblock \emph{Advances in neural information processing systems}, 34: 1273--1286.

\bibitem[{Kallenberg et~al.(2023)Kallenberg, Overweg, van Bree, and Athanasiadis}]{kallenberg2023nitrogen}
Kallenberg, M.~G.; Overweg, H.; van Bree, R.; and Athanasiadis, I.~N. 2023.
\newblock Nitrogen management with reinforcement learning and crop growth models.
\newblock \emph{Environmental Data Science}, 2: e34.

\bibitem[{Kingma and Ba(2014)}]{kingma2014adam}
Kingma, D.~P.; and Ba, J. 2014.
\newblock Adam: A method for stochastic optimization.
\newblock \emph{arXiv preprint arXiv:1412.6980}.

\bibitem[{Lai et~al.(2024)Lai, Wu, Chen, Zhou, and Hovakimyan}]{lai2024residual}
Lai, Z.; Wu, J.; Chen, S.; Zhou, Y.; and Hovakimyan, N. 2024.
\newblock Residual-based Language Models are Free Boosters for Biomedical Imaging Tasks.
\newblock In \emph{Proceedings of the IEEE/CVF Conference on Computer Vision and Pattern Recognition}, 5086--5096.

\bibitem[{Lai, Zhang, and Chen(2024)}]{lai2024adaptive}
Lai, Z.; Zhang, X.; and Chen, S. 2024.
\newblock Adaptive ensembles of fine-tuned transformers for llm-generated text detection.
\newblock \emph{arXiv preprint arXiv:2403.13335}.

\bibitem[{Lee et~al.(2022)Lee, Nachum, Yang, Lee, Freeman, Guadarrama, Fischer, Xu, Jang, Michalewski et~al.}]{lee2022multi}
Lee, K.-H.; Nachum, O.; Yang, M.~S.; Lee, L.; Freeman, D.; Guadarrama, S.; Fischer, I.; Xu, W.; Jang, E.; Michalewski, H.; et~al. 2022.
\newblock Multi-game decision transformers.
\newblock \emph{Advances in Neural Information Processing Systems}, 35: 27921--27936.

\bibitem[{Li et~al.(2023{\natexlab{a}})Li, Wang, Funakoshi, and Okumura}]{li-etal-2023-joyful}
Li, D.; Wang, Y.; Funakoshi, K.; and Okumura, M. 2023{\natexlab{a}}.
\newblock Joyful: Joint Modality Fusion and Graph Contrastive Learning for Multimoda Emotion Recognition.
\newblock In \emph{Proceedings of the 2023 Conference on Empirical Methods in Natural Language Processing}, 16051--16069. Singapore: Association for Computational Linguistics.

\bibitem[{Li et~al.(2022)Li, Puig, Paxton, Du, Wang, Fan, Chen, Huang, Aky{\"u}rek, Anandkumar et~al.}]{li2022pre}
Li, S.; Puig, X.; Paxton, C.; Du, Y.; Wang, C.; Fan, L.; Chen, T.; Huang, D.-A.; Aky{\"u}rek, E.; Anandkumar, A.; et~al. 2022.
\newblock Pre-trained language models for interactive decision-making.
\newblock \emph{Advances in Neural Information Processing Systems}, 35: 31199--31212.

\bibitem[{Li et~al.(2024)Li, Sun, Lai, Wu, Qiu, Xie, Miyata, and Li}]{li2024ecnet}
Li, S.; Sun, K.; Lai, Z.; Wu, X.; Qiu, F.; Xie, H.; Miyata, K.; and Li, H. 2024.
\newblock Ecnet: Effective controllable text-to-image diffusion models.
\newblock \emph{arXiv preprint arXiv:2403.18417}.

\bibitem[{Li et~al.(2023{\natexlab{b}})Li, Fan, Hu, Feichtenhofer, and He}]{li2023scaling}
Li, Y.; Fan, H.; Hu, R.; Feichtenhofer, C.; and He, K. 2023{\natexlab{b}}.
\newblock Scaling language-image pre-training via masking.
\newblock In \emph{Proceedings of the IEEE/CVF Conference on Computer Vision and Pattern Recognition}, 23390--23400.

\bibitem[{Liang et~al.(2023)Liang, Huang, Xia, Xu, Hausman, Ichter, Florence, and Zeng}]{liang2023code}
Liang, J.; Huang, W.; Xia, F.; Xu, P.; Hausman, K.; Ichter, B.; Florence, P.; and Zeng, A. 2023.
\newblock Code as policies: Language model programs for embodied control.
\newblock In \emph{2023 IEEE International Conference on Robotics and Automation (ICRA)}, 9493--9500. IEEE.

\bibitem[{Liu, Jiang, and Pister(2024)}]{liu2024llmeasyquant}
Liu, D.; Jiang, M.; and Pister, K. 2024.
\newblock LLMEasyQuant--An Easy to Use Toolkit for LLM Quantization.
\newblock \emph{arXiv preprint arXiv:2406.19657}.

\bibitem[{Liu et~al.(2024{\natexlab{a}})Liu, Waleffe, Jiang, and Venkataraman}]{liu2024graphsnapshot}
Liu, D.; Waleffe, R.; Jiang, M.; and Venkataraman, S. 2024{\natexlab{a}}.
\newblock GraphSnapShot: Graph Machine Learning Acceleration with Fast Storage and Retrieval.
\newblock \emph{arXiv preprint arXiv:2406.17918}.

\bibitem[{Liu et~al.(2024{\natexlab{b}})Liu, Wu, Bao, Wang, Hovakimyan, and Healey}]{liu2024towards}
Liu, S.; Wu, J.; Bao, J.; Wang, W.; Hovakimyan, N.; and Healey, C.~G. 2024{\natexlab{b}}.
\newblock Towards a Robust Retrieval-Based Summarization System.
\newblock \emph{arXiv preprint arXiv:2403.19889}.

\bibitem[{Liu, He, and Huang(2024)}]{liu2024timemattersexaminetemporal}
Liu, W.; He, Z.; and Huang, X. 2024.
\newblock Time Matters: Examine Temporal Effects on Biomedical Language Models.
\newblock arXiv:2407.17638.

\bibitem[{Madondo et~al.(2023)Madondo, Azmat, Dipietro, Horesh, Jacobs, Bawa, Srinivasan, and O'Donncha}]{madondo2023swat}
Madondo, M.; Azmat, M.; Dipietro, K.; Horesh, R.; Jacobs, M.; Bawa, A.; Srinivasan, R.; and O'Donncha, F. 2023.
\newblock A SWAT-based Reinforcement Learning Framework for Crop Management.
\newblock \emph{arXiv preprint arXiv:2302.04988}.

\bibitem[{Malik, Isla, and Dechmi(2019)}]{malik2019dssat}
Malik, W.; Isla, R.; and Dechmi, F. 2019.
\newblock DSSAT-CERES-maize modelling to improve irrigation and nitrogen management practices under Mediterranean conditions.
\newblock \emph{Agricultural Water Management}, 213: 298--308.

\bibitem[{Mandrini et~al.(2022)Mandrini, Pittelkow, Archontoulis, Kanter, and Martin}]{mandrini2022exploring}
Mandrini, G.; Pittelkow, C.~M.; Archontoulis, S.; Kanter, D.; and Martin, N.~F. 2022.
\newblock Exploring Trade-Offs Between Profit, Yield, and the Environmental Footprint of Potential Nitrogen Fertilizer Regulations in the US Midwest.
\newblock \emph{Frontiers in plant science}, 13.

\bibitem[{Mees, Borja-Diaz, and Burgard(2023)}]{mees2023grounding}
Mees, O.; Borja-Diaz, J.; and Burgard, W. 2023.
\newblock Grounding language with visual affordances over unstructured data.
\newblock In \emph{2023 IEEE International Conference on Robotics and Automation (ICRA)}, 11576--11582. IEEE.

\bibitem[{Mnih et~al.(2015)Mnih, Kavukcuoglu, Silver et~al.}]{mnih2015human-deepRL}
Mnih, V.; Kavukcuoglu, K.; Silver, D.; et~al. 2015.
\newblock Human-level control through deep reinforcement learning.
\newblock \emph{Nature}, 518(7540): 529--533.

\bibitem[{Overweg, Berghuijs, and Athanasiadis(2021)}]{overweg2021cropgym}
Overweg, H.; Berghuijs, H.~N.; and Athanasiadis, I.~N. 2021.
\newblock CropGym: a reinforcement learning environment for crop management.
\newblock \emph{arXiv preprint arXiv:2104.04326}.

\bibitem[{Peng et~al.(2018)Peng, Andrychowicz, Zaremba, and Abbeel}]{peng2018sim2real}
Peng, X.~B.; Andrychowicz, M.; Zaremba, W.; and Abbeel, P. 2018.
\newblock Sim-to-real transfer of robotic control with dynamics randomization.
\newblock In \emph{IEEE International Conference on Robotics and Automation (ICRA)}, 3803--3810.

\bibitem[{Radford et~al.(2018)Radford, Narasimhan, Salimans, Sutskever et~al.}]{radford2018improving}
Radford, A.; Narasimhan, K.; Salimans, T.; Sutskever, I.; et~al. 2018.
\newblock Improving language understanding by generative pre-training.
\newblock \emph{mikecaptain.com}.

\bibitem[{Radford et~al.(2019)Radford, Wu, Child, Luan, Amodei, Sutskever et~al.}]{radford2019language}
Radford, A.; Wu, J.; Child, R.; Luan, D.; Amodei, D.; Sutskever, I.; et~al. 2019.
\newblock Language models are unsupervised multitask learners.
\newblock \emph{OpenAI blog}, 1(8): 9.

\bibitem[{Raman et~al.(2022)Raman, Cohen, Rosen, Idrees, Paulius, and Tellex}]{raman2022planning}
Raman, S.~S.; Cohen, V.; Rosen, E.; Idrees, I.; Paulius, D.; and Tellex, S. 2022.
\newblock Planning with large language models via corrective re-prompting.
\newblock In \emph{NeurIPS 2022 Foundation Models for Decision Making Workshop}.

\bibitem[{Romain et~al.(2022)Romain, Philippe, Julien, Odalric-Ambrym, David et~al.}]{romain2022gym}
Romain, G.; Philippe, P.; Julien, B.; Odalric-Ambrym, M.; David, E.; et~al. 2022.
\newblock gym-DSSAT: a crop model turned into a Reinforcement Learning environment.
\newblock \emph{arXiv preprint arXiv:2207.03270}.

\bibitem[{Schlichtkrull, De~Cao, and Titov(2020)}]{schlichtkrull2020interpreting}
Schlichtkrull, M.~S.; De~Cao, N.; and Titov, I. 2020.
\newblock Interpreting graph neural networks for NLP with differentiable edge masking.
\newblock \emph{arXiv preprint arXiv:2010.00577}.

\bibitem[{Shi et~al.(2017)Shi, Ling, Xue, Qin, Li, Lai, and Yang}]{shi2017combining}
Shi, J.-Y.; Ling, L.-T.; Xue, F.; Qin, Z.-J.; Li, Y.-J.; Lai, Z.-X.; and Yang, T. 2017.
\newblock Combining incremental conductance and firefly algorithm for tracking the global MPP of PV arrays.
\newblock \emph{Journal of Renewable and Sustainable Energy}, 9(2).

\bibitem[{Skhiri and Dechmi(2012)}]{skhiri2012impact}
Skhiri, A.; and Dechmi, F. 2012.
\newblock Impact of sprinkler irrigation management on the Del Reguero river (Spain). I: Water balance and irrigation performance.
\newblock \emph{Agricultural Water Management}, 103: 120--129.

\bibitem[{Sun et~al.(2024{\natexlab{a}})Sun, Ahmed, Ma, Liu, Kabela, Pang, and Kalinli}]{sun2024contextual}
Sun, C.; Ahmed, Z.; Ma, Y.; Liu, Z.; Kabela, L.; Pang, Y.; and Kalinli, O. 2024{\natexlab{a}}.
\newblock Contextual Biasing of Named-Entities with Large Language Models.
\newblock In \emph{ICASSP 2024-2024 IEEE International Conference on Acoustics, Speech and Signal Processing (ICASSP)}, 10151--10155. IEEE.

\bibitem[{Sun et~al.(2017)Sun, Yang, Hu, Porter, Marek, and Hillyer}]{sun2017rl-irrigation}
Sun, L.; Yang, Y.; Hu, J.; Porter, D.; Marek, T.; and Hillyer, C. 2017.
\newblock Reinforcement learning control for water-efficient agricultural irrigation.
\newblock In \emph{2017 IEEE International Symposium on Parallel and Distributed Processing with Applications and 2017 IEEE International Conference on Ubiquitous Computing and Communications (ISPA/IUCC)}, 1334--1341.

\bibitem[{Sun et~al.(2024{\natexlab{b}})Sun, Ren, Li, Wang, and Cao}]{Sun_2024_CVPR}
Sun, S.; Ren, W.; Li, J.; Wang, R.; and Cao, X. 2024{\natexlab{b}}.
\newblock Logit Standardization in Knowledge Distillation.
\newblock In \emph{Proceedings of the IEEE/CVF Conference on Computer Vision and Pattern Recognition (CVPR)}, 15731--15740.

\bibitem[{Tao et~al.(2022)Tao, Zhao, Wu, Martin, Harrison, Ferreira, Kalantari, and Hovakimyan}]{tao2022optimizing}
Tao, R.; Zhao, P.; Wu, J.; Martin, N.~F.; Harrison, M.~T.; Ferreira, C.; Kalantari, Z.; and Hovakimyan, N. 2022.
\newblock Optimizing crop management with reinforcement learning and imitation learning.
\newblock \emph{arXiv preprint arXiv:2209.09991}.

\bibitem[{Wang et~al.(2024)Wang, Wu, Hovakimyan, and Sun}]{wang2024balanced}
Wang, Y.; Wu, J.; Hovakimyan, N.; and Sun, R. 2024.
\newblock Balanced training for sparse gans.
\newblock \emph{Advances in Neural Information Processing Systems}, 36.

\bibitem[{Wright et~al.(2022)Wright, Small, Mackowiak, Grabau, Devkota, and Paula-Moraes}]{wright2022field}
Wright, D.; Small, I.; Mackowiak, C.; Grabau, Z.; Devkota, P.; and Paula-Moraes, S. 2022.
\newblock Field Corn Production Guide: SS-AGR-85/AG202, rev. 8/2022.
\newblock \emph{EDIS}, 2022(4).

\bibitem[{Wu(2024)}]{wu2024exploratory}
Wu, J. 2024.
\newblock \emph{An exploratory journey of representation learning’s enhancement, adaptation and related intelligent methods}.
\newblock Ph.D. thesis, University of Illinois at Urbana-Champaign.

\bibitem[{Wu et~al.(2024{\natexlab{a}})Wu, Chen, Zhao, Sergazinov, Li, Liu, Zhao, Xie, Guo, Ji et~al.}]{wu2024switchtab}
Wu, J.; Chen, S.; Zhao, Q.; Sergazinov, R.; Li, C.; Liu, S.; Zhao, C.; Xie, T.; Guo, H.; Ji, C.; et~al. 2024{\natexlab{a}}.
\newblock Switchtab: Switched autoencoders are effective tabular learners.
\newblock In \emph{Proceedings of the AAAI Conference on Artificial Intelligence}, volume~38, 15924--15933.

\bibitem[{Wu, Hobbs, and Hovakimyan(2023)}]{wu2023hallucination}
Wu, J.; Hobbs, J.; and Hovakimyan, N. 2023.
\newblock Hallucination improves the performance of unsupervised visual representation learning.
\newblock In \emph{Proceedings of the IEEE/CVF International Conference on Computer Vision}, 16132--16143.

\bibitem[{Wu, Hovakimyan, and Hobbs(2023)}]{wu2023genco}
Wu, J.; Hovakimyan, N.; and Hobbs, J. 2023.
\newblock Genco: An auxiliary generator from contrastive learning for enhanced few-shot learning in remote sensing.
\newblock In \emph{ECAI 2023}, 2663--2671. IOS Press.

\bibitem[{Wu et~al.(2024{\natexlab{b}})Wu, Lai, Chen, Tao, Zhao, and Hovakimyan}]{wu2024new}
Wu, J.; Lai, Z.; Chen, S.; Tao, R.; Zhao, P.; and Hovakimyan, N. 2024{\natexlab{b}}.
\newblock The new agronomists: Language models are experts in crop management.
\newblock \emph{arXiv preprint arXiv:2403.19839}.

\bibitem[{Wu et~al.(2023)Wu, Pichler, Marley, Wilson, Hovakimyan, and Hobbs}]{wu2023extended}
Wu, J.; Pichler, D.; Marley, D.; Wilson, D.; Hovakimyan, N.; and Hobbs, J. 2023.
\newblock Extended Agriculture-Vision: An Extension of a Large Aerial Image Dataset for Agricultural Pattern Analysis.
\newblock \emph{arXiv preprint arXiv:2303.02460}.

\bibitem[{Wu et~al.(2022)Wu, Tao, Zhao, Martin, and Hovakimyan}]{wu2022optimizing}
Wu, J.; Tao, R.; Zhao, P.; Martin, N.~F.; and Hovakimyan, N. 2022.
\newblock Optimizing Nitrogen Management with Deep Reinforcement Learning and Crop Simulations.
\newblock In \emph{Proceedings of the IEEE/CVF Conference on Computer Vision and Pattern Recognition Workshops}, 1712--1720.

\bibitem[{Xin et~al.(2024)Xin, Wang, Fu, and Zhou}]{xin2024let}
Xin, W.; Wang, K.; Fu, Z.; and Zhou, L. 2024.
\newblock Let Community Rules Be Reflected in Online Content Moderation.
\newblock \emph{arXiv preprint arXiv:2408.12035}.

\bibitem[{Xu et~al.(2024)Xu, Shang, Feng, Song, Li, Xie, Wang, and Liang}]{xu2024machine}
Xu, W.-J.; Shang, W.-Y.; Feng, J.-M.; Song, X.-Y.; Li, L.-Y.; Xie, X.-P.; Wang, Y.-M.; and Liang, B.-M. 2024.
\newblock Machine learning for accurate detection of small airway dysfunction-related respiratory changes: an observational study.
\newblock \emph{Respiratory Research}, 25(1): 286.

\bibitem[{Yoon et~al.(2020)Yoon, Zhang, Jordon, and Van~der Schaar}]{yoon2020vime}
Yoon, J.; Zhang, Y.; Jordon, J.; and Van~der Schaar, M. 2020.
\newblock Vime: Extending the success of self-and semi-supervised learning to tabular domain.
\newblock \emph{Advances in Neural Information Processing Systems}, 33: 11033--11043.

\bibitem[{Yu et~al.(2024{\natexlab{a}})Yu, Xie, Huang, and Qiu}]{yu2024harnessing}
Yu, C.; Xie, X.; Huang, Y.; and Qiu, C. 2024{\natexlab{a}}.
\newblock Harnessing LLMs for Cross-City OD Flow Prediction.
\newblock \emph{arXiv preprint arXiv:2409.03937}.

\bibitem[{Yu et~al.(2024{\natexlab{b}})Yu, Li, Gao, Liu, and Che}]{yu2024stochastic}
Yu, L.; Li, C.; Gao, L.; Liu, B.; and Che, C. 2024{\natexlab{b}}.
\newblock Stochastic analysis of touch-tone frequency recognition in two-way radio systems for dialed telephone number identification.
\newblock In \emph{2024 7th International Conference on Advanced Algorithms and Control Engineering (ICAACE)}, 1565--1572. IEEE.

\bibitem[{Zhang, Li, and Okumura(2024)}]{zhang2024reconsidering}
Zhang, Y.; Li, D.; and Okumura, M. 2024.
\newblock Reconsidering Token Embeddings with the Definitions for Pre-trained Language Models.
\newblock \emph{arXiv preprint arXiv:2408.01308}.

\bibitem[{Zhang et~al.(2020)Zhang, Wang, Gao, and Liu}]{zhang2020manipulator}
Zhang, Y.; Wang, X.; Gao, L.; and Liu, Z. 2020.
\newblock Manipulator control system based on machine vision.
\newblock In \emph{International Conference on Applications and Techniques in Cyber Intelligence ATCI 2019: Applications and Techniques in Cyber Intelligence 7}, 906--916. Springer.

\bibitem[{Zhao, Queralta, and Westerlund(2020)}]{zhao2020sim}
Zhao, W.; Queralta, J.~P.; and Westerlund, T. 2020.
\newblock Sim-to-real transfer in deep reinforcement learning for robotics: a survey.
\newblock In \emph{2020 IEEE symposium series on computational intelligence (SSCI)}, 737--744. IEEE.

\end{thebibliography}

\end{document}